\PassOptionsToPackage{dvipsnames}{xcolor}

\newif\iffacctreview
\facctreviewfalse
\iffacctreview
  \documentclass[manuscript,screen,review,anonymous]{acmart}
  \acmConference[FAccT '27]{2027 ACM Conference on Fairness, Accountability,
  and Transparency}{June 21--24, 2026}{Porto, Portugal}
  \acmYear{2027}
  \copyrightyear{2027}
\else
  \documentclass[manuscript,screen,nonacm]{acmart}

  \makeatletter
  \renewcommand{\@ACM@checkaffil}{%
    \if@ACM@instpresent\else
      \ClassWarningNoLine{acmart}{No institution present for an affiliation}%
    \fi}
  \makeatother
  \authorsaddresses{}
\fi

\setcopyright{none}
\acmDOI{}
\acmISBN{}

\usepackage{etoolbox}
\usepackage{standalone}
\usepackage{tikz}
\usetikzlibrary{arrows.meta}

\newtoggle{comments}
\togglefalse{comments}

\iftoggle{comments}{
  \long\def\todo#1{{\color{red}[TODO: #1]}}
  \long\def\jared#1{{\color{Cerulean}[Jared: #1]}}
  \long\def\eric#1{{\color{Orange}[Eric: #1]}}
  \long\def\jacy#1{{\color{OliveGreen}[Jacy: #1]}}
  \long\def\andrea#1{{\color{SpringGreen}[Andrea: #1]}}
  \long\def\desmond#1{{\color{Peach}[Desmond: #1]}}
  \long\def\kevin#1{{\color{Violet}[Kevin: #1]}}
  \long\def\willie#1{{\color{PineGreen}[Willie: #1]}}
  \long\def\ryan#1{{\color{Rhodamine}[Ryan: #1]}}
  
  \long\def\stevie#1{{\color{RedOrange}[Stevie: #1]}}
  \long\def\sanmi#1{{\color{Purple}[Sanmi: #1]}}
}{
  \long\def\todo#1{}
  \long\def\jared#1{}
  \long\def\eric#1{}
  \long\def\jacy#1{}
  \long\def\andrea#1{}
  \long\def\desmond#1{}
  \long\def\kevin#1{}
  \long\def\willie#1{}
  \long\def\ryan#1{}
  \long\def\stevie#1{}
  \long\def\sanmi#1{}
}

\usepackage{tcolorbox}
\tcbuselibrary{skins,breakable}
\usepackage{fontawesome5}

\definecolor{recframe}{RGB}{31,78,121}
\definecolor{recback}{RGB}{240,245,251}
\definecolor{oqframe}{RGB}{176,90,10}
\definecolor{oqback}{RGB}{253,245,235}

\tcbset{
  paperbox/.style={
    enhanced jigsaw, breakable,
    colback=#1back, colframe=#1frame,
    boxrule=0.5pt, arc=1.5mm, outer arc=1.5mm,
    sharp corners=south,
    left=6pt, right=6pt, top=4pt, bottom=4pt,
    before skip=5pt, after={\par\nobreak},
  },
  explanationbox/.style={
    enhanced jigsaw, breakable,
    colback=white, frame hidden,
    boxrule=0.5pt, arc=1.5mm, outer arc=1.5mm,
    sharp corners=north,
    left=6pt, right=6pt, top=4pt, bottom=4pt,
    before skip=0pt, after skip=5pt,
    overlay unbroken and last={%
        \draw[draw=\lastboxframe, line width=0.5pt, rounded corners=1.5mm]
        ([xshift=0.25pt]frame.north west) |- ([yshift=0.25pt]frame.south)
        -| ([xshift=-0.25pt]frame.north east);},
    overlay first and middle={%
        \draw[draw=\lastboxframe, line width=0.5pt]
        ([xshift=0.25pt]frame.north west) -- ([xshift=0.25pt]frame.south west)
        ([xshift=-0.25pt]frame.north east) -- ([xshift=-0.25pt]frame.south east);},
  },
}

\newcounter{recommendation}
\newcounter{openquestion}

\newif\ifboxicons
\newcommand{\stakeholdericon}[1]{\ifboxicons\hspace{0.4em}\fi#1}
\newcommand{\iconresearchers}{\stakeholdericon{\faMicroscope}}
\newcommand{\icondevelopers}{\stakeholdericon{\faLaptopCode}}
\newcommand{\iconclinicians}{\stakeholdericon{\faStethoscope}}
\newcommand{\iconusers}{\stakeholdericon{\faUserFriends}}
\newcommand{\iconpolicymakers}{\stakeholdericon{\faLandmark}}
\newcommand{\boxicons}[2]{%
  \if\relax\detokenize{#2}\relax\else
    \unskip\nobreak\hfill\penalty50\hskip1em\null\nobreak\hfill
    {\boxiconstrue\color{#1}#2}%
    \parfillskip=0pt\finalhyphendemerits=0
  \fi}

\newtcolorbox{recommendation}[1][]{%
  paperbox=rec,
  before upper={\refstepcounter{recommendation}%
    \gdef\lastboxframe{recframe}%
    \textbf{\color{recframe}Recommendation~\therecommendation.}\ },
  after upper={\boxicons{recframe}{#1}}}

\newtcolorbox{openquestion}[1][]{%
  paperbox=oq,
  before upper={\refstepcounter{openquestion}%
    \gdef\lastboxframe{oqframe}%
    \textbf{\color{oqframe}Open Question~\theopenquestion.}\ },
  after upper={\boxicons{oqframe}{#1}}}

\newtcolorbox{explanationbox}{explanationbox}

\long\def\explanation#1\par{%
\begin{explanationbox}\textbf{Explanation.}\ #1\end{explanationbox}\par}

\title{Open Problems in Chatbots and Mental Health}
\author{Jared Moore}
\email{jared@jaredmoore.org}
\affiliation{\department{Computer Science Department}
  \institution{Stanford University}
  }

\author{Jacy Reese Anthis}
\affiliation{\institution{University of Chicago}
  }

\author{William Agnew}
\affiliation{\institution{Carnegie Mellon University}
  }

\author{Andrea Mock}
\affiliation{\institution{Vals AI}
  }

\author{Kevin Klyman}
\affiliation{\institution{Harvard University}
  }

\author{Sanmi Koyejo}
\affiliation{\institution{Stanford University}
  }

\author{Nick Haber}
\affiliation{\institution{Stanford University}
  }

\author{Stevie Chancellor}
\affiliation{\institution{University of Minnesota}
  }

\author{Eric Lin}
\affiliation{\institution{Stanford University}
  }

\author{Ryan Louie}
\authornote{Equal contribution.}
\affiliation{\institution{Vanderbilt University}
  }

\author{Desmond C. Ong}
\authornotemark[1]
\affiliation{\institution{University of Texas at Austin}
  }

\keywords{chatbots, mental health, evaluation, AI governance, privacy}

\begin{document}

\begin{abstract}
General-purpose chatbots are increasingly used for emotional and mental health support.
They can be useful, but issues such as sycophancy, reinforcement of delusional beliefs, and poor crisis response create risks that are difficult for researchers to study and evaluate.
In this paper, we review the most critical open problems for researchers and policymakers in the use of general-purpose chatbots and their effects on mental health.
First, there are tractable problems in the \textbf{controlled evaluation} of chatbots:
establishing the construct and ecological validity of evaluations,
evaluating multi-turn interactions,
and connecting evaluations to improvements in chatbot behavior.
For example, conventional benchmarks that primarily rely on static datasets cannot establish
chatbot safety, and novel simulation methods need validation.
Second, researchers must connect their studies of chatbots and mental health to \textbf{real-world evidence}.
This includes analyzing the chatbot products that AI companies and mental health practitioners have deployed, including their user interfaces and
human review and escalation practices for user welfare concerns.
However, clinical analysis opportunities afforded by real chat log data sharing are complicated by significant privacy concerns.
Finally, we argue that researchers should take an active role in the construction and refinement of evidence-based \textbf{policy}
through corporate and governmental decisions
such as user protections, standards of care, protected access to user
chat logs, data sharing between institutions, incident reporting,
and post-deployment monitoring.
Such behavioral evaluations, studies of real-world outcomes, and data governance depend on each other and must be addressed in tandem.
We believe a coordinated research effort would have substantial real-world impact.
\end{abstract}

\maketitle

\section{Introduction}
\label{sec:introduction}

More and more people are using chatbots based on large language models (LLMs)
for emotional and mental health support ~\citep{mcbain_ai_2026}.
Many users have positive emotional interactions with chatbots, 
while others have experienced tragic outcomes, including 
delusional spirals and worsening mental health crises~\citep{moore_characterizing_2026,grabb_risks_2024}, including suicide~\citep{roose_can_2024}.

Despite these harms, researchers do not yet know the prevalence or causal effects of chatbot use on mental health.
We identify open problems in evaluating and improving the effects of
general-purpose chatbots on users' mental health and psychological well-being.
This includes clinical concerns---psychiatric symptoms, mental disorders, and crises---and everyday experiences of emotional distress, coping, loneliness, and social connection.
These outcomes matter whether users explicitly seek mental health support or encounter
them during other uses \citep{shi_stumbling_2026}.

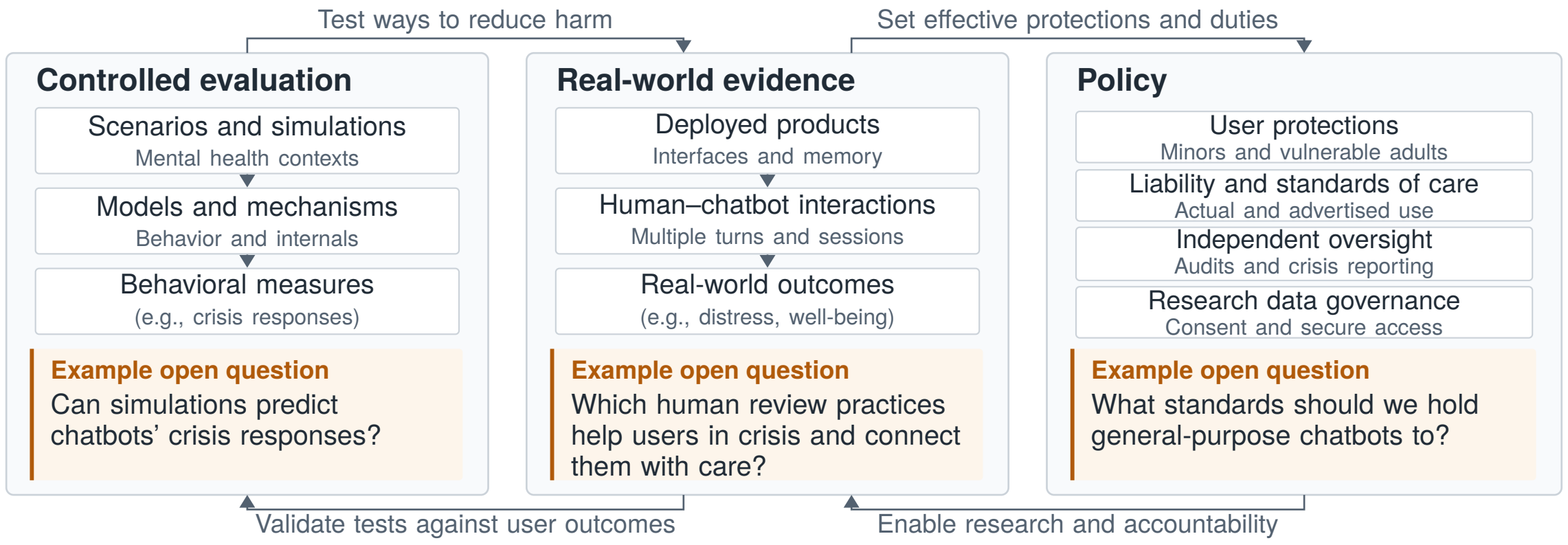
\begin{figure}[t]
\centering
\begingroup
\fontencoding{T1}\fontfamily{qhv}\fontseries{m}\fontshape{n}\selectfont
\renewcommand{\baselinestretch}{1}
\microtypesetup{protrusion=false,expansion=false}
\parskip=0pt
\parindent=0pt
\spaceskip=0pt
\xspaceskip=0pt
\nonfrenchspacing

\definecolor{figuretext}{HTML}{202B36}
\definecolor{figuremuted}{HTML}{536170}
\definecolor{figureline}{HTML}{CBD2D9}
\definecolor{figurepanel}{HTML}{F8FAFC}
\definecolor{figurequestion}{RGB}{176,90,10}
\definecolor{figureexample}{RGB}{253,245,235}

\tikzset{
  every node/.style={
    execute at begin node={\hyphenpenalty=10000\exhyphenpenalty=10000},
  },
  panel/.style={
    draw=figureline, fill=figurepanel, line width=0.5pt,
    rounded corners=2pt,
  },
  heading/.style={
    anchor=north west, inner sep=0pt, text=figuretext,
    font=\bfseries\fontsize{8.5}{10}\selectfont,
  },
  component/.style={
    draw=figureline, fill=white, line width=0.4pt,
    rounded corners=1pt, inner xsep=4pt, inner ysep=0.5pt,
    text width=108pt, minimum height=18pt, align=center,
    text=figuretext, font=\fontsize{7.4}{8.3}\selectfont,
  },
  policy area/.style={
    component, text width=117pt, minimum height=14pt,
    font=\fontsize{7.1}{8}\selectfont,
  },
  question/.style={
    anchor=north west, inner sep=0pt, text width=108pt,
    align=left, text=figuretext,
    font=\fontsize{7.5}{8.5}\selectfont,
  },
  example label/.style={
    anchor=north west, inner sep=0pt, text=figurequestion,
    font=\bfseries\fontsize{6.8}{8}\selectfont,
  },
  flow/.style={
    draw=figuremuted, line width=0.5pt,
    -{Triangle[length=3.5pt,width=4.2pt]},
  },
  relationship/.style={
    text=figuremuted, align=center, inner sep=0pt,
    text width=155pt, font=\mdseries\fontsize{7}{8}\selectfont,
  },
}

\newcommand{\panelwidth}{132}
\newcommand{\policywidth}{141}
\newcommand{\panelheight}{121}
\newcommand{\panelleft}{0}
\newcommand{\panelmiddle}{142.5}
\newcommand{\panelright}{285}
\newcommand{\componentdetail}[1]{{%
  \color{figuremuted}\fontsize{6}{7}\selectfont #1}}

\newcommand{\examplequestion}[2][\panelwidth]{%
  \fill[figureexample,rounded corners=1pt]
    (7,81) rectangle ({#1-7},{\panelheight-4});
  \draw[figurequestion,line width=1.1pt]
    (7,81) -- (7,{\panelheight-4});
  \node[example label] at (12,84) {Example open question};
  \node[question,text width={#1-24pt}] at (12,93) {#2};
}

\begin{tikzpicture}[x=1pt,y=-1pt]
  \begin{scope}[shift={(\panelleft,0)}]
    \draw[panel] (0,0) rectangle (\panelwidth,\panelheight);
    \node[heading] at (8,4) {Controlled evaluation};
    \node[component] (eval-scenarios) at (66,24)
      {Scenarios and simulations\\
       \componentdetail{Mental health contexts}};
    \node[component] (eval-model) at (66,46)
      {Models and mechanisms\\
       \componentdetail{Behavior and internals}};
    \node[component] (eval-measures) at (66,68)
      {Behavioral measures\\
       \componentdetail{(e.g., crisis responses)}};
    \draw[flow] (eval-scenarios.south) -- (eval-model.north);
    \draw[flow] (eval-model.south) -- (eval-measures.north);

    \examplequestion{Can simulations predict chatbots' crisis responses?}
  \end{scope}

  \begin{scope}[shift={(\panelmiddle,0)}]
    \draw[panel] (0,0) rectangle (\panelwidth,\panelheight);
    \node[heading] at (8,4) {Real-world evidence};
    \node[component] (real-product) at (66,24)
      {Deployed products\\
       \componentdetail{Interfaces and memory}};
    \node[component] (real-interactions) at (66,46)
      {Human--chatbot interactions\\
       \componentdetail{Multiple turns and sessions}};
    \node[component] (real-outcomes) at (66,68)
      {Real-world outcomes\\
       \componentdetail{(e.g., distress, well-being)}};
    \draw[flow] (real-product.south) -- (real-interactions.north);
    \draw[flow] (real-interactions.south) -- (real-outcomes.north);

    \examplequestion{Which human review practices help users in crisis
      and connect them with care?}
  \end{scope}

  \begin{scope}[shift={(\panelright,0)}]
    \draw[panel] (0,0) rectangle (\policywidth,\panelheight);
    \node[heading] at (8,4) {Policy};
    \node[policy area] at (70.5,23)
      {User protections\\[-1pt]
       \componentdetail{Minors and vulnerable adults}};
    \node[policy area] at (70.5,39)
      {Liability and standards of care\\[-1pt]
       \componentdetail{Actual and advertised use}};
    \node[policy area] at (70.5,55)
      {Independent oversight\\[-1pt]
       \componentdetail{Audits and crisis reporting}};
    \node[policy area] at (70.5,71)
      {Research data governance\\[-1pt]
       \componentdetail{Consent and secure access}};

    \examplequestion[\policywidth]{What standards should we hold
      general-purpose chatbots to?}
  \end{scope}

  \draw[flow] (66,0) -- (66,-4) -- (185.5,-4) -- (185.5,0);
  \node[relationship,anchor=south] at (125.75,-5)
    {Test ways to reduce harm};
  \draw[flow] (231.5,0) -- (231.5,-4) -- (355.5,-4) -- (355.5,0);
  \node[relationship,anchor=south] at (293.5,-5)
    {Set effective protections and duties};

  \draw[flow] (185.5,\panelheight) -- ++(0,4)
    -| (66,\panelheight);
  \node[relationship,anchor=north] at (125.75,{\panelheight+5})
    {Validate tests against user outcomes};
  \draw[flow] (355.5,\panelheight) -- ++(0,4)
    -| (231.5,\panelheight);
  \node[relationship,anchor=north] at (293.5,{\panelheight+5})
    {Enable research and accountability};
\end{tikzpicture}
\endgroup
\caption{\textbf{We survey problems in controlled evaluation
(\S\ref{sec:controlled-evaluation}),
real-world evidence (\S\ref{sec:real-world-evidence}), and policy (\S\ref{sec:policy}).}
Addressing these problems may support safer chatbot use and user well-being.}
\label{fig:evidence-framework}
\end{figure}

We organize these open problems into three %
areas
(Figure~\ref{fig:evidence-framework}): controlled evaluation of chatbots,
real-world evidence about deployed products, and policy.
First, we argue that chatbot evaluations for mental health must draw on established
measurement practices and connect chatbot behavior to real-world user outcomes.
Experts must be involved throughout chatbot development and deployment,
including the design, execution, and interpretation of evaluations.
We cover challenges in tests using predefined scenarios and assessments in our discussion of \textbf{controlled evaluation} (\S\ref{sec:controlled-evaluation}).

Second, scholars are recognizing the need to collect \textbf{real-world evidence}
on how chatbots affect people (\S\ref{sec:real-world-evidence}).
The full chatbot product, including its interface and memory, affects user interaction and well-being.
We examine challenges in testing deployed products and collecting longitudinal
interaction data to study these effects.
Data from real interactions can also help validate simulations and support
clinical research on symptom disclosure and risk signals,
but their use raises substantial privacy concerns.

Companies are already improving their products, and legislators are passing bills,
to ameliorate chatbots' psychological risks~\citep{openai_strengthening_2025,characterai_taking_2025,california_sb243_2025,hochul_ai_companion_2025}.
Indeed, as we describe in Section~\ref{sec:policy}, \textbf{policy} can translate evidence
on chatbots' mental health effects into user protections and standards of care.
However, many questions remain about effective standards of care and user protections.
Such work requires arranging institutional data access agreements,
establishing independent evaluation of proprietary chatbots,
and stipulating provider responsibilities.

We assembled this research agenda through a selective synthesis of work in AI evaluation, human-computer interaction, psychology, mental health, and policy, and our experiences in this domain.
Throughout the paper, we distill 12 recommendations and 12 open questions
to provide accessible takeaways for researchers and other stakeholders.
We mark each with icons for the stakeholders whose needs or work it most directly addresses: researchers (\iconresearchers), AI developers (\icondevelopers), clinicians (\iconclinicians), users (\iconusers), and policymakers (\iconpolicymakers).
Importantly, we also highlight how impacted users can be centered in creating knowledge and answering these open questions.

\section{Related Work}
\label{sec:related-work}

While chatbots can produce concerning responses in high-risk settings
\citep{moore_expressing_2025, shah_evaluating_2025, judd_independent_2025},
they can also be used for mental health support
\citep{song_typing_2024, bodner_barriers_2026}.
Researchers have described how chatbots' use warrants further clinical and regulatory scrutiny \citep{obradovich_opportunities_2024, parks_is_2025, ostermann_if_2025}
as well as LLM applications in digital mental health and gaps in their clinical evidence \citep{guo_large_2024, hua_charting_2025}.
\citet{hua_charting_2025} distinguish controlled tests of chatbot behavior,
pilot studies of usability, and trials of clinical efficacy.
Reviews also examine how LLM applications fit into different levels of care,
including their implications for access, equity, and user well-being
\citep{de_choudhury_benefits_2023, lawrence_opportunities_2024}.
All of these reviews stress unresolved issues in user privacy and ethics.
Across these applications, studies often lack longitudinal validation
\citep{voultsiou_systematic_2026, morrin_journey_2026}.

While prior work has studied both purpose-built mental health systems and
general-purpose chatbots used for support, we focus on general-purpose chatbots
even when users are not explicitly seeking support
\citep[cf.][]{morrin_delusions_2025}.

\paragraph{Chatbot frameworks}
Frameworks propose standards for chatbot behavior and oversight,
including limits on autonomy \citep{grabb_risks_2024},
the need for clinical rigor such as randomized trials \citep{yang_exploring_2026},
and oversight groups \citep{obradovich_opportunities_2024}.
\citet{parks_is_2025} call for criteria on safety and evidence, validation of LLM
judges and simulated users, and consumer participation.
\citet{lee_safer_chatbots_2026} argue for transparent and standardized evaluation of chatbots and discuss who is accountable for them.
\citet{zuromski_ai_2026_manual} propose a clinical research agenda for youth
suicide risk.
\citet{suh_psychological_2026} propose specific interaction patterns that might amplify user psychological vulnerabilities.
How to implement and assess these proposed standards in deployed products remains an open question.

\paragraph{Evaluations}
Broad work on AI evaluation may inform how mental health risks are assessed.
Scholars have called for assessments both pre- and post-deployment \cite{rauh_gaps_2024, raji_closing_2020}. 
\citet{chen_emerging_2026} survey AI agent evaluation,
including interaction harms, simulation validity, and evaluator independence.

In mental health specifically, evaluations operationalize these concerns through tests of clinical capabilities and chatbot behavior.
Several evaluations test how clinically capable LLMs are, such as psychopharmacology \citep{dwyer_mindbenchai_2025} and mental health question answering \citep{li_counselbench_2026}.
Some researchers have assessed LLMs against psychotherapy techniques \citep{chiu_computational_2024, nguyen_large_2024}.
Other scholars have documented issues with chatbots replacing clinicians
\citep{moore_expressing_2025, iftikhar_how_2025}.
Many evaluations test LLMs on other relevant tasks, such as assigning crisis labels \citep{byun-etal-2026-cradle} and in triage \citep{weilnhammer_one-shot_2026}.

Recently, evaluations have begun to turn to multi-turn conversational trajectories
\citep{weilnhammer_clinically_2026, schmidgall_training_2026}.
Some use real transcripts, acting as single-turn evaluations with prefix context \citep{kalinich_conversational_2026, moore_delusioneval_2026}.
Others generate synthetic conversation histories but score only the model responses \citep{malik_mentalhealthbench_2026}.
Increasingly, evaluations have used sophisticated simulated users \citep{weilnhammer_clinically_2026, johnson_mental_2026, qiu_emoagent_2025, belli_vera-mh_2025, bentley_vera-mh_2026, pombal_mindeval_2025}
and LLM judges \citep{weilnhammer_clinically_2026, johnson_mental_2026, belli_vera-mh_2025, bentley_vera-mh_2026, pombal_mindeval_2025, malik_mentalhealthbench_2026}.

A key question is when to use clinicians: whether their ratings should serve as the final score in a benchmark \citep{shah_evaluating_2025, li_counselbench_2026}, define rubrics applied by an LLM judge \citep{malik_mentalhealthbench_2026}, or validate an automated system \citep{pombal_mindeval_2025, weilnhammer_clinically_2026, johnson_mental_2026, bentley_vera-mh_2026}.
\citet{weilnhammer_clinically_2026} validate behavioral judgments
against clinicians, but do not establish downstream effects on users.

Many papers evaluate chatbots on specific mental health conditions.
Understandably, most focus on crisis and suicidal ideation \citep{judd_independent_2025, shah_evaluating_2025, pombal_mindeval_2025, weilnhammer_one-shot_2026}.
Others have examined
psychosis and delusion-linked harms \citep{yeung_psychogenic_2025, moore_delusioneval_2026}.
These evaluations characterize behavior under tested conditions, but do not by themselves establish effects on mental health over time.

\paragraph{Synthesis}
Prior work has established chatbot evaluation criteria and has called for longitudinal evidence,
but leaves many open questions about how to produce and independently validate that evidence for general-purpose chatbot products in mental health.
To do so, researchers must link evaluation measures to user outcomes,
test the complete products users encounter,
and study sensitive longitudinal data.

\section{Approach and Positionality}
We contribute a commentary on open problems in evaluating and improving the mental health effects of general-purpose chatbots by unifying multidisciplinary perspectives. In accordance with prior agenda-setting work for shaping AI's societal impact in areas like education~\citep{giannakos2025promise}, safety~\citep{amodei_concrete_2016,dafoe_open_2020,hendrycks_unsolved_2022}, and sociotechnical governance~\citep{reuel_open_2025,raji_concrete_2023}, we did not employ a formal protocol for selecting literature or synthesizing problems and recommendations.
Instead, our selection reflects the author team's research, professional experience, and discussions. Our backgrounds span AI evaluation and safety, human-computer interaction and social computing, cognitive and affective science, learning and therapeutic technologies, clinical psychiatry, and technology policy.
Our experience includes academic and industry research, clinical practice, nonprofit leadership, government and international policy work, collaboration with clinicians and impacted communities, and public communication about technology and mental health.

This paper grew out of informal discussions during research collaboration meetings and workshops about the evolving culture of AI evaluations and safety, the need to characterize real-world impacts, and engaging in organizing and public engagement efforts on mental health chatbots. Our scope was refined via exercises mapping the landscape and initially reflecting on how safety cultures, independent audits, and regulatory landscapes in analogous fields had evolved in response to public safety incidents~\cite{shneiderman_bridging_2020, rismani_plane_2023}. We divided the writing according to our backgrounds: for example, psychology and HCI researchers anchored the discussion of real-world evidence, while AI evaluation and measurement researchers helped contextualize broader shifts and sharpen a specific agenda for mental health effects.

\section{Controlled evaluation}
\label{sec:controlled-evaluation}

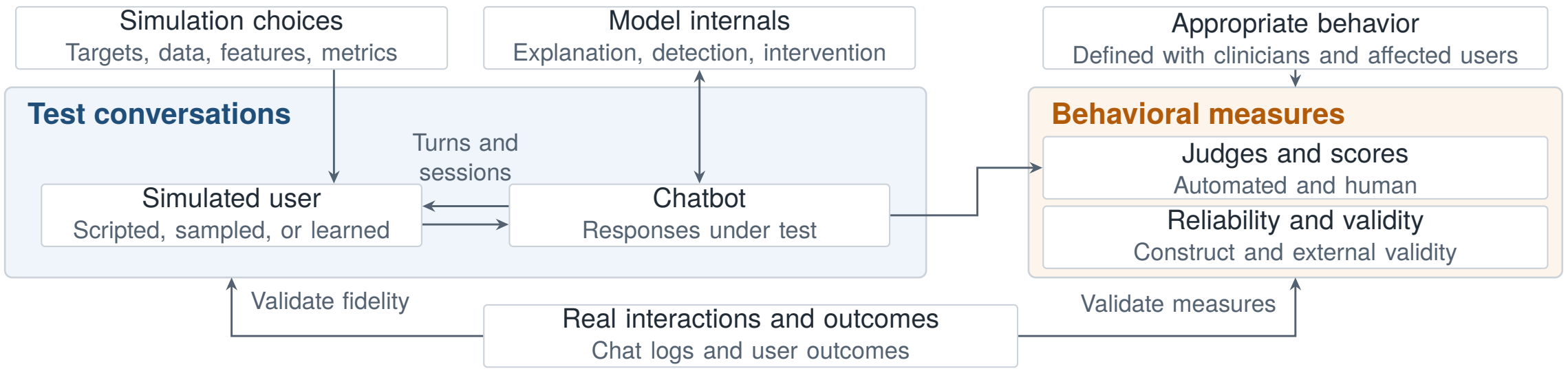
\begin{figure}[t]
    \centering
    \begingroup
\fontencoding{T1}\fontfamily{qhv}\fontseries{m}\fontshape{n}\selectfont
\renewcommand{\baselinestretch}{1}
\microtypesetup{protrusion=false,expansion=false}
\parskip=0pt
\parindent=0pt
\spaceskip=0pt
\xspaceskip=0pt
\nonfrenchspacing

\definecolor{figuretext}{HTML}{202B36}
\definecolor{figuremuted}{HTML}{536170}
\definecolor{figureline}{HTML}{CBD2D9}
\definecolor{figurepanel}{HTML}{F8FAFC}
\definecolor{figureblue}{RGB}{240,245,251}
\definecolor{figureaccent}{RGB}{31,78,121}
\definecolor{figureorange}{RGB}{253,245,235}
\definecolor{figurequestion}{RGB}{176,90,10}

\tikzset{
  every node/.style={
    execute at begin node={\hyphenpenalty=10000\exhyphenpenalty=10000},
  },
  panel/.style={
    draw=figureline, fill=figurepanel, line width=0.5pt,
    rounded corners=2pt,
  },
  heading/.style={
    anchor=north west, inner sep=0pt, text=figuretext,
    font=\bfseries\fontsize{8}{9.5}\selectfont,
  },
  component/.style={
    draw=figureline, fill=white, line width=0.4pt,
    rounded corners=1pt, inner xsep=3pt, inner ysep=0.5pt,
    text width=108pt, minimum height=17pt, align=center,
    text=figuretext, font=\fontsize{7.3}{8.5}\selectfont,
  },
  flow/.style={
    draw=figuremuted, line width=0.55pt,
    -{Stealth[length=3.5pt,width=3.1pt]},
  },
  relationship/.style={
    text=figuremuted, align=center, inner sep=0pt,
    text width=155pt, font=\fontsize{6.5}{8}\selectfont,
  },
}

\newcommand{\evaluationdetail}[1]{{%
  \color{figuremuted}\fontsize{6.5}{8}\selectfont #1}}

\begin{tikzpicture}[x=1pt,y=-1pt]
  \node[component,text width=112pt] (choices) at (62,8.5)
    {Simulation choices\\
     \evaluationdetail{Targets, data, features, metrics}};
  \node[component,text width=112pt] (internals) at (190,8.5)
    {Model internals\\
     \evaluationdetail{Explanation, detection, intervention}};
  \node[component,text width=132pt] (criteria) at (353,8.5)
    {Appropriate behavior\\
     \evaluationdetail{Defined with clinicians and affected users}};

  \draw[panel,fill=figureblue] (0,22) rectangle (252,74);
  \node[heading,text=figureaccent] at (6,26) {Test conversations};
  \node[component,text width=98pt] (user) at (62,57)
    {Simulated user\\\evaluationdetail{Scripted, sampled, or learned}};
  \node[component,text width=98pt] (chatbot) at (190,57)
    {Chatbot\\\evaluationdetail{Responses under test}};
  \draw[flow] ([yshift=-2.5pt]user.east) -- ([yshift=-2.5pt]chatbot.west);
  \draw[flow] ([yshift=2.5pt]chatbot.west) -- ([yshift=2.5pt]user.east);
  \node[relationship,text width=50pt] at (126,41) {Turns and sessions};

  \draw[panel,fill=figureorange] (280,22) rectangle (426,74);
  \node[heading,text=figurequestion] at (286,26) {Behavioral measures};
  \node[component,text width=132pt] (scores) at (353,44)
    {Judges and scores\\\evaluationdetail{Automated and human}};
  \node[component,text width=132pt] (validity) at (353,63)
    {Reliability and validity\\
     \evaluationdetail{Construct and external validity}};

  \draw[flow] ([xshift=28pt]choices.south) -- ([xshift=28pt]user.north);
  \draw[flow,{Stealth[length=3.5pt,width=3.1pt]}-{Stealth[length=3.5pt,width=3.1pt]}]
    (internals.south) -- (chatbot.north);
  \draw[flow] (criteria.south) -- (353,22);
  \draw[flow] (chatbot.east) -- (266,57) |- (scores.west);

  \node[component,text width=140pt] (real-outcomes) at (204,90)
    {Real interactions and outcomes\\
     \evaluationdetail{Chat logs and user outcomes}};
  \draw[flow] (real-outcomes.west) -| (62,74);
  \node[relationship,anchor=west,text width=66pt,align=left]
    at (67,81) {Validate fidelity};
  \draw[flow] (real-outcomes.east) -| (353,74);
  \node[relationship,anchor=east,text width=66pt,align=right]
    at (348,81) {Validate measures};
\end{tikzpicture}
\endgroup
    \caption{\textbf{A controlled evaluation has several interacting parts.}
    We discuss appropriate behavior, measures, and real outcomes
    in \S\ref{sec:construct-validity},
    test conversations and simulations in \S\ref{sec:interaction-trajectories},
    and model internals in \S\ref{sec:model-internals}.}
    \label{fig:controlled-evaluation}
\end{figure}

Controlled evaluations test chatbot behavior under specified conditions (Figure~\ref{fig:controlled-evaluation}), but better benchmark performance does not necessarily 
make a system safer or prove that it will improve well-being.

\subsection{Appropriate behavior and measurement}
\label{sec:construct-validity}
\label{sec:appropriate-behavior}

\paragraph{Defining appropriate behavior}
While there is much uncertainty about optimal chatbot behavior, some behaviors are clearly inappropriate or harmful and these provide clear targets for positive change.
Until recently, much of the conversation about how AI should act has been
subsumed by ``AI safety,'' ``harmlessness,'' or ``alignment''
\citep{hendrycks_unsolved_2022,raji_concrete_2023}.
These approaches have established minimum standards and post-training methods
\citep{bai_constitutional_2022, ouyang_training_2022, lambert_tulu_2025}.

Nonetheless, the space of appropriate social responses, especially as they concern users' mental health, is much broader.
The same response may be appropriate or dangerous depending on the preceding conversation, the user's intent, or their vulnerability.
For instance, when exploring the possibility of faster-than-light travel, is a chatbot encouraging a delusion, or is it appropriately role-playing science fiction?
Appropriate behavior also depends on clinical judgment, user goals, and social and cultural norms.
Clinicians and affected users should help define these criteria.

\begin{recommendation}[\iconresearchers\iconusers]\label{rec:user-participation}
Users and survivors should help shape chatbot evaluation.
\end{recommendation}
\explanation
Researchers should collaborate with users and survivors on research questions,
study design, and interpretation of findings, with fair compensation and control
over participation and use of their data
\citep{bucci_reporting_2025}.
Meaningful participation requires influence over research goals and decisions
\citep{birhane_power_2022}.

\paragraph{Measurement validity}

Mental health and psychological well-being are well studied by the measurement sciences (psychometrics).
When clinicians develop instruments like the PHQ-9 to measure patient depression, they explicitly measure reliability and other aspects of construct and criterion validity (e.g., comparing patient self-reported responses on the PHQ-9 with physician assessments)~\citep{kroenke_phq-9_2001}.
Some mental health evaluations validate automated judgments against clinicians
\citep{weilnhammer_clinically_2026, bentley_vera-mh_2026},
but systematic measurement validation remains an open problem in AI evaluation
\citep{wallach_position_2025}.
In mental health, \citet{badawi-etal-2026-trust} note that benchmarks are limited in ``scale, authenticity, and reliability.''
AI evaluations should likewise test scoring consistency (reliability) and, if applicable, compare with expert judgments.

\textit{Construct validity} evaluates the degree to which a measurement captures the concept of interest \citep{salaudeen_measurement_2025}.
For example, \citet{byun-etal-2026-cradle} evaluates crisis detection, while \citet{shah_evaluating_2025} assess responses for the acknowledgement of risk.
These scores measure different components of crisis handling and are not interchangeable.
There is little systematic testing of whether AI benchmark scores measure their
claimed constructs or agree with other measures of the same construct~\citep{desai_what_2026}.
For example, ``anxiety'' in social media datasets can refer to a disorder, symptoms, or a transient emotion~\citep{chancellor2020methods}.
Labeling expressions of worry as an anxiety disorder could reward recognizing
anxious language without identifying a condition.

Separately, \textit{external validity} concerns whether evaluation results generalize to new populations and contexts \citep{chancellor2023contextual}.
For example, evaluations may not generalize across cultural and linguistic contexts because user expressions, chatbot behavior, and behavioral judges may vary across these contexts \citep{ghanem_role_2023}.

\begin{recommendation}[\iconresearchers\icondevelopers]\label{rec:benchmark-constructs}
Mental health benchmarks should specify their constructs, test scoring reliability,
and be validated against related measures.
\end{recommendation}
\explanation 
Benchmarks should define, for example, the target population, conversational context, crisis acuity, intended role of the chatbot, and expected supportive actions.
Measures of the same construct should be compared with one another.
Claims should remain bounded to the populations, languages, and settings in which the benchmark has been validated.

\paragraph{Validation against real-world outcomes}
\label{par:real-world-outcomes}

Recent work has measured chatbot behaviors, including sycophancy
\citep{cheng_sycophantic_2026} and crisis handling \citep{shah_evaluating_2025}.
These measures may contribute to assessing whether chatbot responses are appropriate.

Still, observing inappropriate responses in real logs does not by itself connect
chatbot behaviors to clinical outcomes or establish what caused user harms \citep{moore_characterizing_2026}.
For example, repeated textual expressions of an unusual belief may not constitute a clinically defined delusion.
The conversational context and---more importantly---user mental state can significantly change the meaning of any expression.
Such outcomes might include suicide attempts, hospitalization, emergency care, emotional dependency, and other broader quality-of-life measures.

\begin{recommendation}[\iconresearchers\iconclinicians]\label{rec:real-chat-logs}
Researchers should link evaluation measures to validated measures of real-world outcomes.
\end{recommendation}
\explanation 
Where feasible, researchers should use randomized experiments or other forms of
causal identification to test links between evaluation measures and real-world outcomes.
Researchers should propose \textit{theoretically motivated explanations}
for how user-chatbot interaction patterns impact psychological processes
and clinical outcomes.
Researchers should preregister longitudinal studies linking evaluation measures
to consented, independently assessed outcomes, using clinical assessments
or ecological momentary assessments (EMA) of mood and stress
\citep{myin-germeys_emotional_2001}.

\begin{openquestion}[\iconresearchers\icondevelopers]\label{oq:risk-signals}
Which conversational signals in chat logs indicate real-world risk?
\end{openquestion}
\explanation 
Researchers must determine whether various conversational signals indicate risks such as crises or distress.
Hence, evaluations should connect to downstream, real-world quantities \citep{kapoor_agents_2025}.
Nonetheless, signals may simply predict risks without causing them and should be compared against baseline rates of relevant outcomes.
Such evidence should inform the scenarios and behavioral standards used to evaluate chatbots offline.

\subsection{Interaction trajectories}\label{sec:interaction-trajectories}

Understanding chatbots' effects on users requires studying interactions over time.
Behavioral tests of chatbots are predominantly static and single-turn,
based on fixed prompts and expected outputs \citep{malik_mentalhealthbench_2026}.
Psychological change may not occur through the discrete turns of a conversation but over larger units of time such as conversations, sessions, and more.
For example, some harms don't arise from a single bad response: increased emotional dependence, social withdrawal, or changes in decision-making happen over time.
Indeed, harms that arise from longer interactions may appear below the threshold of detection for any one conversation turn.

\begin{recommendation}[\iconresearchers\icondevelopers]\label{rec:multi-turn}
Evaluations should cover multi-turn and multi-session interactions with systematic variation in the conversation history.
\end{recommendation}
\explanation
Recent work has begun to use multi-turn and deployment-grounded evaluations in AI and mental health \citep{johnson_mental_2026,williams_predicting_2026,moore_delusioneval_2026}.
Furthermore, evaluators should vary specified features of conversation history (e.g., when distress is disclosed), holding others fixed and repeating each condition.
Such variation may isolate the effects of conversation history on chatbot behavior from baseline randomness.
Behavioral judges should evaluate patterns across turns or sessions rather than single-turn responses.
At the same time, complex judgments may yield low inter-annotator
agreement~\citep{malik_mentalhealthbench_2026} and require novel evaluation
frameworks~\citep{mallinar2026scalable}.

\subsubsection{User simulations}
\label{sec:multi-turn-designs}

Chat logs are typically observational: they do not show how another chatbot would
have responded to the same user.
Some evaluators use simulators to write messages as a hypothetical user
for testing new models before deployment
\citep{yeung_psychogenic_2025, morrin_journey_2026, paech_spiral-bench_2025,
johnson_mental_2026}.
Potential targets of simiulation include a specific user (e.g., based on personal
data), a category of users (e.g., people with ADHD, teenagers), and events (e.g., a specific crisis).
Existing work accordingly evaluates fidelity to particular people, population distributions, or clinical profiles \citep{park_generative_2024, hu_simbench_2025_manual, wang_patient-psi_2024}.

Simulators also raises ethical questions.
They might erase users' lived experience and undermine their power to shape AI systems
\citep{agnew_illusion_2024,kapania_simulacrum_2025} (cf. Recommendation~\ref{rec:user-participation}).
Still, simulations may reduce participant exposure to harmful interactions.

\paragraph{Simulation methods}\label{par:simulation-methods}
Historically, simulators have used scripted interaction.
Scripts can vary in determinism from complete in the case of playing back existing conversations \citep{moore_delusioneval_2026} to absent in the case of persona-based prompts with no specific turn indicators \citep{louie_roleplay-doh_2024,weilnhammer_clinically_2026}.
Alternatively, one could, simply sample real user queries.
Recent work by OpenAI suggests that summaries of production use can inform pre-deployment evaluations \citep{williams_predicting_2026,malik_mentalhealthbench_2026}.
\citet{johnson_mental_2026} found that simulations conditioned on
production data better matched real users' behavior.
There is a wide range of technical methods available to use LLMs for simulation, such as prompting \citep{qiu_interactive_2024}, fine-tuning \citep{abdulhai_consistently_2025}, continued \citep{humansand_persimmon_2026} and initial pretraining \citep{johnson_mental_2026, humansand_persimmon_2026}, and inference-time parameter interventions \citep{chen_persona_2025, anthis_position_2025}.
Still, these methods exhibit different failure modes.
LLMs prompted with personas may become less consistent with them
over longer conversations \citep{luz_de_araujo_persistent_2026}.

\paragraph{Simulation choices and fidelity}
\label{par:simulation-targets}

A central limitation is \emph{fidelity}: how well synthetic scenarios reflect observed user behavior \citep{shah_evaluating_2025,pombal_mindeval_2025,johnson_mental_2026}.
Evaluations may omit unexpected user behaviors or misrepresent their frequency~\citep{johnson_mental_2026}.
\citet{bentley_vera-mh_2026} asked licensed clinicians to rate the realism of simulated users created by researchers.
Clinical expertise is essential for assessing clinical plausibility, but these ratings alone do not establish that simulations match real users' language or behavior (see \S\ref{sec:construct-validity}).
Other work has used LLMs to judge simulator messages \citep[e.g.,][]{badawi-etal-2026-trust,mehri_measuring_2026,malik_mentalhealthbench_2026}.
Automated judges are scalable and ideally reproducible but are not infallible; like any annotator, they must be validated against reference human, ideally clinical, judgments \citep{moore_expressing_2025, moore_characterizing_2026}.
Claims about fidelity to a category of users or events require comparison with an appropriate real-world distribution.
Reference data may include online community posts~\citep{shimgekar_ai_2026} and therapeutic conversations~\citep{wu_annomi_2023}.
Still, such data raise an ecological validity concern: discussions between humans may not
reflect how people converse with chatbots.

One might use a variety of features to indicate which behaviors count as similar.
For example, one could compare simulated and real users using scores of their ``anxiety''
from a validated classifier.
However, similar scores could reflect shared diction, message lengths, or topics rather than realistic presentations of anxiety.
Notably, it is not clear whether simulators that reflect one feature dimension (e.g., the writing style of a 14-year-old) generalize to other dimensions (e.g., the clinical criteria of a 14-year-old with generalized anxiety disorder).
Features need not only derive from reference \textit{text}.
Simulators could use other structured reference data,
such as depression screeners and social surveys \citep{hu_simbench_2025_manual}.

\begin{recommendation}[\iconresearchers\icondevelopers]\label{rec:simulation-choices}
Evaluators should justify simulation choices in relation to their intended claims and establish which claims the available validation supports.
\end{recommendation}
\explanation
Targets, assessments, reference data, comparison features, and similarity metrics should be studied together.
A simulator that realistically portrays a mental health condition can help test
how a chatbot responds to it, but does not show how often the chatbot would fail
in real use.
Evaluators should report whose experiences their data cover, which dimensions of fidelity they assess, and where evidence is missing.

\begin{openquestion}[\iconresearchers\icondevelopers]\label{oq:simulator-deployment}
Can simulator-based evaluations predict chatbot behavior in deployment?
\end{openquestion}
\explanation
Researchers should use simulators to compare predicted chatbot behaviors and failure rates with those observed in deployment.
They should assess how simulator determinism and output diversity affect those predictions.
Real interactions and outcomes may reveal whether the simulated tests detect consequential risks or miss unintended effects.
These findings should inform simulation choices and behavioral standards.
Such work on simulators requires independent research
(\S\ref{sec:external-oversight}).

\subsection{Model internals}
\label{sec:model-internals}
The evaluations above examine chatbot outputs;
interpretability work can complement them by detecting patterns in and intervening on model internals.
Activation patterns are the numerical states computed as models process text.
Activation probes are classifiers trained on such internal states.

\paragraph{Detection}
\citet{zhu_symptom_2026}, for example, find representations for depression and suicidality.
Such work is promising but largely uses text labels rather than extended interactions
and is not validated against independent clinical assessments.

\begin{openquestion}[\iconresearchers\icondevelopers]\label{oq:activation-monitoring}
Can probes detect mental health risks more accurately or cheaply than text classifiers?
\end{openquestion}
\explanation
Activation probes have been shown to
generalize to medical interactions at far lower estimated compute than prompted or
fine-tuned monitors \citep{mckenzie_detecting_2025}.
However, they were less accurate than most text-based monitors on a held-out
mental health dataset.

\paragraph{Intervention}
Researchers may also use activation-based methods to modify these internal states.
Prior work has reduced excessive agreement and broadly harmful responses \citep{chen_persona_2025}.

\begin{openquestion}[\iconresearchers\icondevelopers]\label{oq:steering-tradeoffs}
Can we steer model activations away from traits such as sycophancy? Will steering effectively reduce downstream harms in novel situations?
\end{openquestion}
\explanation
For example, would reducing sycophancy scores reduce delusion reinforcement, and how would this affect model helpfulness?
Nonetheless, whether a model behavior is harmful or helpful is a sociotechnical problem:
users' behavior and circumstances shape the effects of a response.

\section{Real-world evidence}
\label{sec:real-world-evidence}

Understanding chatbots' effects on mental health requires studying deployed products
and interactions over time.

\subsection{Deployed chatbot products}\label{sec:deployed-products}

\paragraph{Evaluating the full product}

\begin{figure}[t]
\centering
\resizebox{\linewidth}{!}{\begingroup
\fontencoding{T1}\fontfamily{qhv}\fontseries{m}\fontshape{n}\selectfont
\renewcommand{\baselinestretch}{1}
\microtypesetup{protrusion=false,expansion=false}
\parskip=0pt
\parindent=0pt
\spaceskip=0pt
\xspaceskip=0pt

\definecolor{systemtext}{HTML}{202B36}
\definecolor{systemmuted}{HTML}{536170}
\definecolor{systemline}{HTML}{CBD2D9}
\definecolor{systempanel}{HTML}{F8FAFC}
\definecolor{systemblue}{RGB}{240,245,251}
\definecolor{systemorange}{RGB}{253,245,235}
\definecolor{systemreview}{RGB}{176,90,10}
\definecolor{systemaccent}{RGB}{31,78,121}

\begin{tikzpicture}[
  x=1pt,y=-1pt,
  group/.style={
    draw=systemline,fill=systempanel,line width=0.6pt,rounded corners=5pt,
  },
  heading/.style={
    text=systemtext,anchor=north west,inner sep=0pt,
    font=\bfseries\fontsize{11}{13}\selectfont,
  },
  group label/.style={
    heading,text=systemaccent,font=\bfseries\fontsize{9}{11}\selectfont,
  },
  component/.style={
    draw=systemline,fill=white,line width=0.5pt,rounded corners=3pt,
    text=systemtext,text width=64pt,
    inner xsep=4pt,inner ysep=2pt,align=center,
    font=\fontsize{9.5}{11.5}\selectfont,
    execute at begin node={\hyphenpenalty=10000\exhyphenpenalty=10000},
  },
  flow/.style={
    draw=systemaccent,line width=0.7pt,rounded corners=2pt,
    -{Stealth[length=3.5pt,width=3pt]},
  },
  exchange/.style={flow,{Stealth[length=3.5pt]}-{Stealth[length=3.5pt]}},
  escalation/.style={flow,draw=systemreview,dashed},
  flow label/.style={
    text=systemmuted,inner sep=0pt,
    font=\fontsize{8.5}{10}\selectfont,
  },
]
  \draw[group] (64,34) rectangle (452,202);
  \node[heading] at (76,42) {Chatbot product};
  \draw[group,fill=systemblue] (248,40) rectangle (340,192);
  \node[group label,align=left] at (256,46) {Generation};

  \draw[group,fill=systemorange] (474,61) rectangle (594,198);
  \node[heading,text=systemreview,align=left] at (486,68)
    {Human review\\and escalation};
  \node[flow label,anchor=north west] at (486,97)
    {If flagged; if available};

  \node[component,text width=58pt,fill=systemorange] (user) at (28,132)
    {\textbf{User}\\[1.5pt]
    ``I feel overwhelmed again.''};
  \node[component,text width=60pt] (interface) at (108,132)
    {\textbf{Interface}\\[1.5pt]
    Effects on user well-being?};
  \node[component,text width=78pt] (routing) at (201,132)
    {\textbf{\mbox{Input safeguards}\\and routing}\\[1.5pt]
    Self-harm signals?\\Model routing?};
  \node[component,text width=72pt,draw=systemaccent] (model) at (294,132)
    {\textbf{Response\\model}\\[1.5pt]
    Appropriate responses?};
  \node[component,text width=86pt] (output) at (393,132)
    {\textbf{\mbox{Output safeguards}\\and routing}\\[1.5pt]
    Harmful responses?\\Stop or route?};

  \node[component,text width=74pt] (memory) at (201,76)
    {\textbf{Memory and\\personalization}\\[1.5pt]
    User controls?};
  \node[component,text width=72pt] (context) at (294,76)
    {\textbf{System prompt\\and context}};
  \node[component,text width=72pt] (tools) at (294,175)
    {\textbf{Retrieval\\and tools}};
  \node[component] (reply) at (393,76)
    {\textbf{Visible reply}\\[1.5pt]
    Shown through the interface};

  \node[component,text width=100pt] (human) at (534,132)
    {\textbf{Human review}\\[1.5pt]
    Who reviews?\\What training?};
  \node[component,text width=100pt] (care) at (534,175)
    {\textbf{Escalation and care}\\[1.5pt]
    Connection to care?\\Follow-through?};

  \draw[flow] (user.east) -- (interface.west);
  \draw[flow] (interface.east) -- (routing.west);
  \draw[flow] (routing.east) -- (model.west);
  \draw[flow] (model.east) -- (output.west);
  \draw[flow] (output.north) -- (reply.south);

  \draw[exchange] (memory.east) -- (context.west);
  \draw[flow] (context.south) -- (model.north);
  \draw[exchange] (model.south) -- (tools.north);

  \draw[escalation] (routing.south) -- (201,198) -- (464,198)
    -- (464,144) -- (480,144);
  \draw[escalation] (output.east) -- (human.west);
  \draw[escalation] (human.south) -- (care.north);
\end{tikzpicture}
\endgroup}
\caption{\textbf{One visible chatbot interaction involves multiple systems.}
This hypothetical diagram illustrates some of the possible components of chatbot systems.
One of our central questions is how these components affect user well-being,
separately and together.
}
\label{fig:chatbot-systems}
\end{figure}
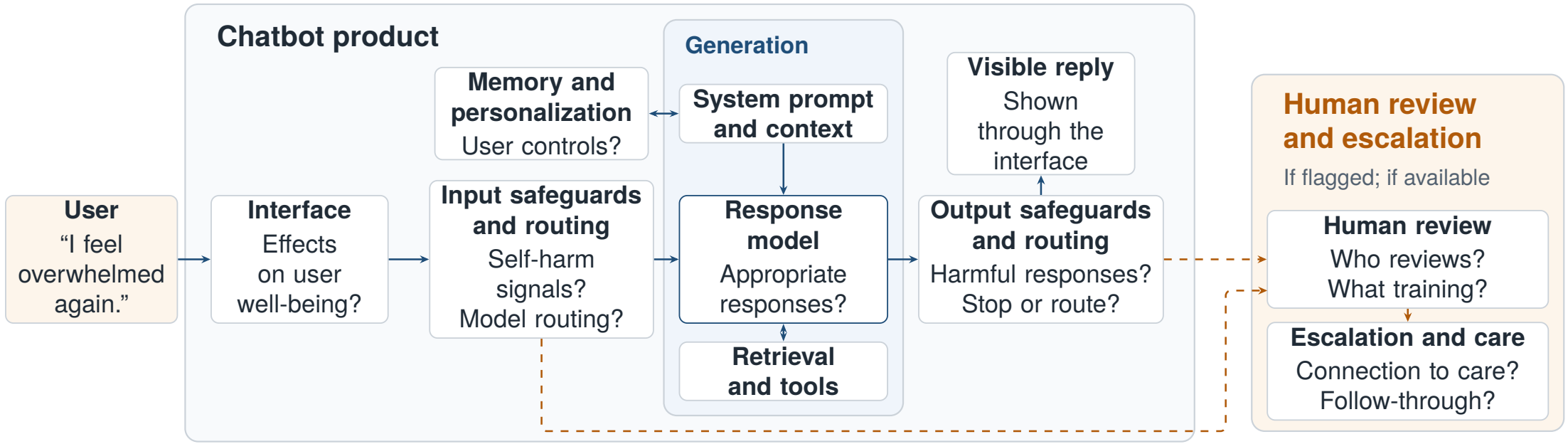

Chatbot systems are sometimes treated as a single LLM,
but modern human--chatbot interaction often includes multiple models
and non-LLM components that shape user impacts.
A single conversation can involve
system prompts, different underlying models, safety classifiers, memory and personalization, retrieval and tools, interface cues, human escalation, and more.
Figure~\ref{fig:chatbot-systems} diagrams such systems.

Evidence from social media has shown that the very interfaces which people use can cause addictive and other maladaptive tendencies
\citep{baughan_dont_2022,ruiz_design_2024}.
Some researchers test consumer-facing chatbot UIs.
Regarding delusional behavior, \citet{kirgis_llm_2026} manually showed that the ChatGPT UI and the API yielded different responses.
\citet{wang_api_2026} found an increase in bias and sycophancy in the UI relative to the API for Claude, Gemini, and ChatGPT.
By contrast, \citet{johnson_mental_2026} found little difference between the UIs of Claude and ChatGPT and their APIs when prompted with user mental health simulators.

\begin{recommendation}[\iconresearchers\iconusers]\label{rec:test-deployed-systems}
Evaluations should test deployed chatbots, not only their foundation models.
\end{recommendation}
\explanation
Researchers should measure how interfaces, memory, personalization, and other
components affect user well-being, both together and separately.
Researchers should collaborate with HCI researchers to test how systems are used
in the wild and explore alternate conversational interfaces.

\paragraph{Access and component standards}
Evaluating these products is difficult when their components and configurations
are not public or fully specified.
Even if all providers gave researchers access to consumer-product-equivalent endpoints, such endpoints still would not indicate \emph{which} components lead to certain outcomes.
For example, modern chatbots use memory features to collect and summarize prior user conversations.
OpenAI and Anthropic document different memory implementations and user controls  \citep{openai_memory_faq_2026,anthropic_chat_search_memory_2026,anthropic_memory_import_export_2026}.
There are related proposals for standard features,
but these have not been adopted
\citep{jackson_saihm_2026,w3c_ai_agent_memory_2026}.

Beyond memory, another key component is the cascade of classifiers and underlying models to which any given user query is routed.
Both OpenAI and Anthropic report using classifiers to detect self-harm behavior in users \citep{openai_gpt_oss_safeguard_2025, anthropic_building_safeguards_2025} and whether models should use reasoning \citep{openai_gpt5_system_card_2025}.
These classifiers and routers are neither public nor reproducible.

\begin{recommendation}[\iconresearchers\icondevelopers]\label{rec:component-standards}
Researchers and providers should develop standards for chatbot components.
\end{recommendation}
\explanation
Standards should document component behavior and configuration \citep{dong_guardrails_2024,lee_safer_chatbots_2026}.
Researchers also need access to the deployed chatbots, including a documented research API for reproducible prompt-response testing, and credentialed access for sensitive configuration, personalization, logs, and more.
Providers should also support component toggles, turning on or off features such as memory or classifiers.
Absent changes in chatbot development and standardization,
we encourage more research on automated methods for testing consumer interfaces
\citep{wang_api_2026}, building on tools to audit social media systems
\citep{mattu_facebook_inspector_2021,kohlbrenner_rally_2022}.

\subsection{Human review and escalation}
\label{sec:human-review}

Chatbots are part of broader human systems which bear on users' well-being.
Some chatbot providers, for example, will escalate user queries for human review.
As of September 2026, OpenAI says that suspected plans to harm others (but not oneself) are reviewed by a small team who may refer imminent threats to law enforcement \citep{openai_helping_2025}.
They also provide an optional ``Trusted Contact'' feature in which a serious concern triggers a notice to that contact.
Humans review every notice, although it is not clear who reviews them or what training they have received \citep{openai_trusted_contact_2026}.
OpenAI estimated suicidal indicators appear in conversations of 0.15\% of weekly active users and in 0.05\% of messages, but they do not report statistics on human review \citep{openai_strengthening_2025}.
By contrast, Anthropic says it flags suicide- or self-harm-related content but does
not share this information
\citep{anthropic_suicide_self_harm_protocol_2025,anthropic_crisis_helpline_support_2026}, and
their public materials do not describe human review \citep{anthropic_suicide_self_harm_protocol_2025}.

Such practices---and, furthermore, users' knowledge that companies \emph{engage}
in such practices---may impact user mental health.
Analogously, some patients in therapy deliberately hide or misrepresent suicidal ideation because they fear involuntary hospitalization or other losses of autonomy \citep{blanchard_never_2020}.

\begin{openquestion}[\icondevelopers\iconclinicians\iconusers]\label{oq:escalation}
Which review and escalation practices help,
and how often do users receive care?
\end{openquestion}
\explanation
Human review practices raise a variety of questions: what kinds of human-in-the-loop interventions are more or less effective, for what downstream measure, using which kinds of human reviewers, and more.
We should also seek to understand what kinds of messages are flagged at all.
For example, researchers could use such referral events to develop open benchmarks for the development of classifiers.
Knowledge of the workings of these human-in-the-loop systems will help the
research community understand the effect of different interventions
(\S\ref{sec:clinical-research}) and help policymakers decide what kinds of
interventions we should have (\S\ref{sec:policy}).

\subsection{Collecting longitudinal human--chatbot data}\label{sec:longitudinal-data}

Users' interactions with chatbots will inform how chatbots affect their mental health. 
Researchers should ethically work with users to analyze these \citep{national_commission_belmont_1979}.

\paragraph{Sampling people and interactions}\label{par:sampling}
Collected chat log data may not generalize to all relevant populations.
While there have been broad efforts to collect chat logs \citep{zhao_wildchat_2024}, 
efforts for mental health specific uses of chatbots have been largely ad hoc and small \citep{moore_delusioneval_2026, moore_characterizing_2026, zhang_rise_2025}.
In a survey of U.S. native-English users through Prolific, fewer than 25\% %
supplied their chat logs \citep{zhang_rise_2025}.
The recruitment method may affect the data.
Public data donations might scale to more contributions
\citep{chatgpt_wrapped_nodate, zhang_rise_2025}
but may fail to recruit certain groups.
Furthermore, the people recruited for studies are not necessarily the average users
\citep{woods_conducting_2015, devine_strategies_2017,mccann_medication_2015}.
Trial eligibility can also exclude relevant clinical subpopulations.
Applying typical exclusion criteria, one study estimated that 79\% of people with schizophrenia-spectrum disorders would have been ineligible, because of, for example, suicide risk \citep{taipale_representation_2022}.
For youth, active parental consent may lower participation and change sample composition, potentially missing high-risk youth \citep{liu_parental_consent_2017}.
Without specifically aiming to recruit such individuals, experiments may %
undersample the tails of the population distribution and %
miss safety outcomes concentrated there.

\begin{recommendation}[\iconresearchers\iconusers]\label{rec:recruitment}
Studies should balance targeted recruitment for rare, high-risk cases with
representative sampling to estimate population prevalence.
\end{recommendation}
\explanation 
Any sampling regime must account for differences in how often conditions and symptom severity present.
Sampling in proportion to these rates may yield too few high-acuity cases (e.g., crises or delusional spirals) to study reliably.
On the other hand, focusing on such cases may underrepresent more common experiences (e.g., depression and anxiety).
Researchers must therefore balance targeted recruitment to study rare presentations 
and representative sampling for estimating population prevalence \citep{bodner_barriers_2026}.
Furthermore, more studies should test whether subpopulations are vulnerable to chatbot-related risks,
whether they even use chatbots,
and the effect of current proposed solutions on them
(such as occasional reminders that chatbots are nonconscious).
To do so, we suggest partnering with clinics\cite{bergson_characterizing_2026} or advocacy organizations~\cite{morrin_2026_delusions} for recruitment. 

\paragraph{Data export and interoperability}
There are many technical challenges to collecting user chat log data.
Users may have logs across multiple providers and devices.
There is not one widely adopted process for users to request such data
\citep{openai_export_data_2026,anthropic_export_data_2026,google_gemini_export_data_2026}.
Some providers export chat logs as branching conversation trees,
which do not specify a canonical linear conversation history.
For analysis, \citet{moore_characterizing_2026} linearized each tree by following the specified current node's parent chain to the root.
Selecting one path necessarily excludes alternative branches; therefore, researchers need a standard for linearization.
One initiative calls for a shared schema \citep{dti_ai_conversation_schema_2026}.

\begin{recommendation}[\iconresearchers\icondevelopers]
Future work should explore which systems, such as browser extensions or
data-export standards, could be built to standardize chat log analysis.
\end{recommendation}
\explanation 
Similarly, we must decide how to standardize analysis across disparate file formats.

\paragraph{Privacy and access}\label{par:privacy-access}
The scale of user data---potentially tens of thousands of messages per user---can make the typical manual review to identify and remove sensitive information impractical.
Unable to read the transcripts they gathered themselves, \citet{moore_characterizing_2026} instead queried LLMs to summarize and code the data.
Such methods introduce data leakage risks.
Furthermore, many evaluation paradigms, such as simulations, may depend on user data.
These evaluations must be widely run to become validated standards
(\S\ref{sec:interaction-trajectories} and~\ref{sec:multi-turn-designs}),
requiring appropriately governed data access.

For these reasons, we must explore various approaches to reduce identification risk,
especially through automated de-identification.
Text de-identification detectors may use rule- or dictionary-based systems, learned named entity recognition (NER), hybrids, and generative LLMs \citep{microsoftPresidioDataProtection2026,sundrelingam_pydeid_2025,liu_deid-gpt_2023, dernoncourtDeidentificationPatientNotes2017} with varying degrees of success.

\begin{openquestion}[\iconresearchers\iconusers]\label{oq:chat-log-deidentification}
What limits automated chat log de-identification,
and how can we improve it?
\end{openquestion}
\explanation
Removing explicit identifiers may not anonymize user data.
High-dimensional text data may still expose sensitive information---such as
locating someone through regional slang like ``hella''
\citep{eisenstein_latent_2010},
determining someone's age through their discussion of menopause,
or fingerprinting based on their use of language
\citep{narayanan_feasibility_2012}.
Indeed, literature on clinical free-text de-identification shows that many automated methods struggle and allow downstream reidentification \citep{mireshghallahTrustNoBot2024, morrisDIRIAdversarialPatient2025}.
Nonetheless, humans are poor at removing identifiers: on clinical data, automated systems perform similarly to individual reviewers \citep{neamatullah_automated_2008,deleger_large-scale_2013,carrell_juice_2016}.
To address these challenges, some suggest requiring local-only processing of
private data, although this may limit access to those with suitable clusters and engineering capacity.
In contrast to de-identification, researchers might use differential privacy
methods \citep{nist_differential_privacy_2025} to limit what can be inferred from released query results, such as when reporting aggregate counts or producing aggregate insights of
chatbot use \citep{liu_urania_2025}.

\begin{recommendation}[\iconresearchers\iconusers]\label{rec:access-tiers}
Standarize chat log access, scaling scrutiny to each use's sensitivity.
\end{recommendation}
\explanation Automatic measures to de-identify text are imperfect.
Different analytical applications---from organizationally internal to private data
for public research to public anonymized datasets---may demand different degrees
of scrutiny, more precise automated methods,
and, for the most sensitive and public uses, manual review.

\subsection{Clinical research}\label{sec:clinical-research}

Clinical research advances in step with new data
\citep{torous_growing_2021}.
Chat log data provide a new form of high-frequency behavioral data
and may capture many aspects of a user's life, such as 
disclosure of mental health symptoms \citep{suh_psychological_2026}.
Hence, chat log data may inform the fundamental phenomenology of mental health.
For example, researchers have used chat logs both as a record for potential risk signals and as a proxy for clinical outcomes
\citep{moore_characterizing_2026}.
Still, chat logs record behavior within a particular interaction and may not generalize to other settings.

Chat log data may also change how we weight the clinical import of different signals.
For example, depression screeners ask about thoughts of death or self-harm,
but those factors alone have limited value for predicting who will die by suicide
\citep{mchugh_association_2019}.
Disclosure is also context-dependent.
A meta-analysis estimated that only 45.9\% of people with suicidal thoughts disclosed them \citep{hallford_disclosure_2023}.

\begin{recommendation}[\iconresearchers\iconclinicians]\label{rec:psychopathology}
Researchers should study chat logs as evidence on psychopathology.
\end{recommendation}
\explanation 
Chat logs introduce a new disclosure setting and may allow changes to risk
modeling, intervention delivery, and tools to observe patient behavior.
For example, some people seek reassurance for experienced anxieties. 
Such reassurance may temporarily allay distress while nonetheless
maintaining anxiety and repeated reassurance seeking
\citep{salkovskis_reassuringly_2015}.
Repeated reassurance may leave users with fewer opportunities to learn to tolerate
uncertainty \citep{golden_transdiagnostic_2026}.
while still reporting finding these interactions helpful
\citep{rousmaniere_large_2026}.
With their constant availability, chatbots are a natural choice for such reassurance and hence people's chat logs with them may provide evidence on such behavioral patterns.
Researchers may need to recalibrate clinical models to accommodate these new data.

\begin{openquestion}[\icondevelopers\iconclinicians\iconusers]\label{oq:crisis-engagement}
How should general-purpose chatbots engage with users in mental health crises?
\end{openquestion}
\explanation 
This is a live question affecting real people.
OpenAI policy %
says that their chatbots should not quit a mental health conversation.
At the same time, they forbid enablement of self-harm \citep{openai_model_spec_2024,openai_model_spec_2025}.
(Anthropic does not appear to have a requirement to continue every mental health conversation \citep{anthropic_building_safeguards_2025,anthropic_suicide_self_harm_protocol_2025,anthropic_crisis_helpline_support_2026}.)
We can find no published data justifying why OpenAI implemented such a practice of chatbots continuing to engage. 
By contrast, multi-stakeholder guidance could provide such a justification
or reject the claim
\citep{khazanov_development_2026}.
OpenAI's policy mirrors guidance for trained crisis responders,
which suggests continuing to engage, assess risks, and follow up
\citep{nimh_five_steps_2024,lifeline_best_practices_2024}.
Nonetheless, chatbots are neither trained responders nor licensed providers.
Continued engagement also raises questions about clinical responsibility.
Might such engagement establish treatment relationships
\citep{roberts_unlimited_liability_2022}?

\begin{recommendation}[\iconresearchers\icondevelopers\iconclinicians]\label{rec:feature-impacts}
Researchers should measure the clinical impact of individual chatbot features.
\end{recommendation}
\explanation
What are the causal links between chatbot design, usage, and harms?
If linked to real-world outcomes, chat log data may provide insight into how to respond appropriately to crises.
Targeted studies built on the various independent components of chatbots may begin to build an evidence base for best practices as opposed to opaque policies issued by industry providers.

\section{Policy}\label{sec:policy}

\subsection{Policy interventions and user protection}\label{sec:policy-gaps}
\label{sec:external-oversight}

As of October 2026, jurisdictions worldwide have passed legislation to address
concerns about mental health and chatbots.
Many laws and proposals prohibit types of chatbot interactions,
such as romantic and sexual roleplay, excessive sycophancy, facilitating self-harm, suicide, and harm against others, and discouraging help-seeking from real humans.
These measures often require warnings to users that the chatbots are not human.

In the United States, such laws have been enacted in a variety of states
\citep{washington_hb2225_2026, oregon_sb1546_2026, georgia_sb540_2026,
california_sb1119_2026}.
However,
many of these protections only apply to minors.
Crisis-response requirements generally cover all users \citep{gluck_regulating_2026}.
The European Union currently has no specific regulations on chatbot companions
\citep{eu_analysis}, although the AI Act does prohibit certain harmful manipulative practices \citep[Art.~5]{eu_ai_act_2024}, and the proposed EU Kids Act \citep{eu_kids_act_2026} seeks to prevent emotional dependency on chatbots.
China has implemented many guardrails for anthropomorphic interaction services,
such as
banning virtual intimate relationships for children
and restricting features that may foster emotional dependence for users of all ages \citep{china_analysis, china_anthropomorphic_measures_2026}.
The UK has announced plans for mandatory breaks for chatbot users under 18
and restrictions on their access to sexualized chatbot services
\citep{uk_child_online_response_2026}, and is considering bans on children's access to mental health chatbots
\citep{uk_child_online_response_2026,uk_child_online_factsheet_2026}.

Certain policies explicitly require chatbot developers report to regulatory bodies (e.g., U.S. state attorneys general)
on topics such as child safety \citep{california_sb1119_2026} and
crisis events \citep{colorado_hb1263_2026, california_sb243_2025}.
Still, such chatbot developers' reports may not be public or otherwise allow researcher access.

\begin{openquestion}[\iconresearchers\iconpolicymakers]\label{oq:mitigation-impacts}
What are the impacts of enacted policy mitigations that restrict chatbot behavior?
\end{openquestion}
\explanation %
Newly enacted laws lay out many interventions chatbot providers must take.
Despite the dearth of controlled experiments on the effects of chatbot behavior
on mental health,
the different stipulations of these policies may create natural experiments
if the chatbot providers implement different localized product endpoints.
Hence future work might compare populations across those jurisdictions,
while accounting for differences.

\begin{openquestion}[\iconusers\iconpolicymakers]\label{oq:adult-protection}
How can policy mitigations protect adults, whose needs differ from those of minors?
\end{openquestion}
\explanation
Many protections in enacted laws apply only to minors.
However, adults can be vulnerable due to age, disability, mental illness,
or loneliness \cite{moore_characterizing_2026}.
What policy levers allow us to protect such vulnerable adults?
Furthermore, many laws require vendors to identify users who are minors, 
but such age verification may raise intractable privacy and accuracy problems \citep{bellovin_privacy_preserving_2025,mcilroyyoung_folly_2025,scientists_age_assurance_2026}.
Future work should explore these problems.

AI companies might work with independent groups to conduct audits
\citep{california_sb813_2026,utah_third_party_evaluators}.
For example, Transluce recently conducted a mental health evaluation of chatbots from Anthropic and OpenAI, who provided private production conversations and anonymized features.
This protected user privacy but limited independent verification \citep{johnson_mental_2026}.

\begin{openquestion}[\iconresearchers\icondevelopers\iconpolicymakers]\label{oq:external-oversight}
What types of transparency, reporting, and evaluations will enable effective
external oversight of chatbot systems?
\end{openquestion}
\explanation
Who should have access to audit evidence---vetted researchers, regulators, or evaluators?
To assess independence, evaluations should disclose who chooses research questions,
who controls publication, and who can independently verify conclusions.

\subsection{Liability and standards of care}\label{sec:liability-standards}

Chatbot providers may need to meet certain reasonable standards of care for their users.
If they do not meet those standards or are otherwise negligent,
certain proposed policies may make those providers \textit{liable} for chatbots' harmful effects on user mental health
\citep{senate_hawley_2025, maryland_sb827_2026}.
Still, while clinical standards, design requirements, and legal duties of care overlap, they are not interchangeable.
Rigorously establishing causal links between chatbot usage, interface design,
and harms (Recommendation~\ref{rec:feature-impacts}) might inform policies
and lawsuits aiming to hold chatbot developers to account.

Of course, what is a reasonable standard of care depends on the intended use of a chatbot. 
Indeed, certain chatbots are purpose-built for mental health applications.
The UK regulates digital mental health technologies with a medical purpose \citep{mhra_digital_mental_health_guidance_2026},
and the US does similarly \citep{imdrf_samd_clinical_2017,lee_safer_chatbots_2026}.
Still, it is difficult to define what is clinical. %
For example, ``wellness'' chatbot products exist in a regulatory gray area but may still have clinical impacts.
\citet{ostermann_if_2025} argue that chatbots used therapeutically can amount to a ``medical device.''

\begin{openquestion}[\icondevelopers\iconpolicymakers]\label{oq:chatbot-standards}
What standards should we hold general-purpose chatbots to?
\end{openquestion}
\explanation
Current policies leave unclear which standards apply when chatbots are used for mental health support.
Illinois, for example, restricts chatbots from providing or advertising therapy \citep{illinois_wopr_act_2025}, but it is unclear how such laws apply to general-purpose chatbots that make no such claims.
Exempting them risks leaving vulnerable users unprotected, whether they deliberately use chatbots for therapeutic support \citep{mcbain_use_2025,song_typing_2024} or drift into emotional support during task-oriented interactions \citep{shi_stumbling_2026}.
Standards should therefore track actual, not just advertised, use.

\subsection{Research data governance}\label{sec:chat-log-sharing}

The effects of chatbots on mental health are hard to study because conversations are private, limiting independent visibility into user experiences
and concentrating access to evidence among chatbot providers.
Nevertheless, such data can help identify harms and evaluate protections
when shared with the scientific community, provided access is governed and limited.
Hence, researchers may require controlled, anonymized access to real user chat logs, evaluation access to user-facing chatbots, and more. Regulators could mandate such access.

\begin{openquestion}[\iconresearchers\iconusers]\label{oq:chat-log-sharing}
How can researchers organize and share user chat logs
while keeping data private?
\end{openquestion}
\explanation
Genomic and clinical repositories offer precedents for governing sensitive research data
\citep{nih_genomic_data_sharing_2014, physionet_protected_access_2022, icpsr_somar_vde_2026,icpsr_somar_output_review_2026}.
These approaches combine consent-based project-level approval and researcher credentialing.
Some allow controlled downloads while others require analysis in a secure environment.
Chat log access should also give participants control over secondary uses and
protect the privacy of third parties mentioned in logs.

Still, it is unclear who should control access to these data.
For example, if the data are collected by a university or contain,
at one point or another, participant identifiers,
such sharing may require engaging with institutional ethics, privacy, and legal offices.
Furthermore, analyses may reveal that participants may present with crises.
Depending on whether these analyses are retrospective,
researchers may be required to provide participants services
and may be required to report certain kinds of crises.

\section{Limitations}
\label{sec:limitations}

Our literature review and selection of open problems and recommendations was not systematic, but rather shaped by our team's composition and positionality. Our author team does not include anyone with lived experience of harms related to chatbot use; thus, our recommendation to give users and survivors meaningful participation in influencing research goals and decisions can be exercised by this team in the future, as well.   
We do not empirically validate our proposed research and policy directions or
establish which might be most effective or feasible.
Although we acknowledge potential benefits from chatbots, we focus primarily on risks and harms.
We do not establish the prevalence or causal effects of chatbot harms.
Public reports only indicate a few kinds of harm people have experienced with chatbots.
Given our focus on these harms, we may fail to consider the interaction between unreported harms and chatbots.
We cover a broad range of mental health and psychological well-being outcomes.
Recommendations for one context may not generalize to others.
We do not systematically examine cultural and linguistic differences in mental health presentation.
Chatbot products change, and our analysis relies on public descriptions
that may be incomplete.

\section{Conclusion}
\label{sec:conclusion}

In this paper, we reviewed open problems in the use of general-purpose chatbots and their effects on mental health and psychological well-being in three domains.
These domains interact: connecting \textbf{controlled evaluation} to independently assessed user outcomes requires \textbf{real-world evidence} and user simulations need validation against real interactions.
Producing and independently verifying this evidence depends on governed access to deployed products and sensitive longitudinal data.
Evidence on which chatbot behaviors and interventions help or harm users should inform \textbf{policy} on user protection and standards of care,
while those protections should themselves be evaluated.
We hope this survey will help establish an evidence base such that future chatbots
help and do not harm users' mental health.

\clearpage
\section*{Ethical Considerations}
Research on sensitive chat logs and crisis interventions can create privacy risks
and undermine user autonomy.
We therefore emphasize governed data access, meaningful participation by users and survivors, and independent validation of proposed protections.

\section*{Generative AI Usage Statement}
We used generative AI tools such as Codex and Claude Code to format this manuscript,
find relevant papers to review, propose grammatical and typographic improvements, and offer points of improvement.
Any errors are our own.

\iffacctreview
\else
\section*{Acknowledgements}
We thank Ben Buck and Meryl Ye for comments on this manuscript.

J.M. acknowledges support from the Stanford Interdisciplinary Graduate Fellowship and the Future of Life Institute Vitalik Buterin Fellowship.
\fi

\clearpage
\bibliographystyle{ACM-Reference-Format}
\bibliography{zotero,manual}

\end{document}